\documentclass[a4paper,twoside]{article}

\usepackage{epsfig}
\usepackage{subfigure}
\usepackage{calc}
\usepackage{amssymb}
\usepackage{amstext}
\usepackage{amsmath}
\usepackage{amsthm}
\usepackage{multicol}
\usepackage{pslatex}
\usepackage{apalike}
\usepackage{SCITEPRESS}     
\usepackage{booktabs}
\usepackage{multirow}
\usepackage{csquotes} 
\usepackage{float}
\usepackage{booktabs}
\usepackage[table,xcdraw]{xcolor}

\newcommand{\leqnomode}{\tagsleft@true\let\veqno\@@leqno}
\usepackage{cuted}

\begin{document}

\title{Improving Bag-of-Visual-Words Towards Effective Facial Expressive Image Classification}

\author{\authorname{Dawood Al Chanti\sup{1} and Alice Caplier\sup{1}}
\affiliation{\sup{1}Univ. Grenoble Alpes, CNRS, Grenoble INP\thanks{Institute of Engineering Univ. Grenoble Alpes}
, GIPSA-lab, 38000 Grenoble, France}
\email{dawood.alchanti@gmail.com}}


\abstract{Bag-of-Visual-Words (BoVW) approach has been widely used in the recent years for image classification purposes. However, the limitations regarding optimal feature selection, clustering technique, the lack of spatial organization of the data and the weighting of visual words are crucial. These factors affect the stability of the model and reduce performance. We propose to develop an algorithm based on BoVW for facial expression analysis which goes beyond those limitations. Thus the visual codebook is built by using k-Means++ method to avoid poor clustering. To exploit reliable low level features, we search for the best feature detector that avoids locating a large number of keypoints which do not contribute to the classification process. Then, we propose to compute the relative conjunction matrix in order to preserve the spatial order of the data by coding the relationships among visual words. In addition, a weighting scheme that reflects how important a visual word is with respect to a given image is introduced. We speed up the learning process by using histogram intersection kernel by Support Vector Machine to learn a discriminative classifier. The efficiency of the proposed algorithm is compared with standard bag of visual words method and with bag of visual words method with spatial pyramid. Extensive experiments on the CK+, the MMI and the JAFFE databases show good average recognition rates. Likewise, the ability to recognize spontaneous and non-basic expressive states is investigated using the DynEmo database.}

\onecolumn \maketitle \normalsize \vfill

\section{\uppercase{Introduction \& Prior Art}}
\label{sec:introduction}

\noindent Bag of Visual Words model (BoVW) with distinctive local features generated around keypoints has become the most popular method for image classification tasks. It has been first introduced by \cite{sivic2003video} for object matching in videos. Sivic and Zisserman described the BoVW method as an analogy with text retrieval and analysis. Wherein, a document is represented by word frequencies without regard to their order. The word frequencies are considered as the signature of the document and are then used to perform classification. 

Since 2003 till nowadays, it obtained state-of-the-art performance on several applications in computer vision such as human action recognition \cite{peng2016bag}, scene classification \cite{zhu2016bag} and face recognition \cite{hariri2017geometrical}. To the best of our knowledge, this approach has not been investigated for the task of facial expression recognition since around 2013 \cite{ionescu2013local}. In this paper, we aim at introducing some improvements over the standard BoVW method to tackle facial expression recognition.

\begin{figure*}[]
\centering
\includegraphics[width=5.0 in]{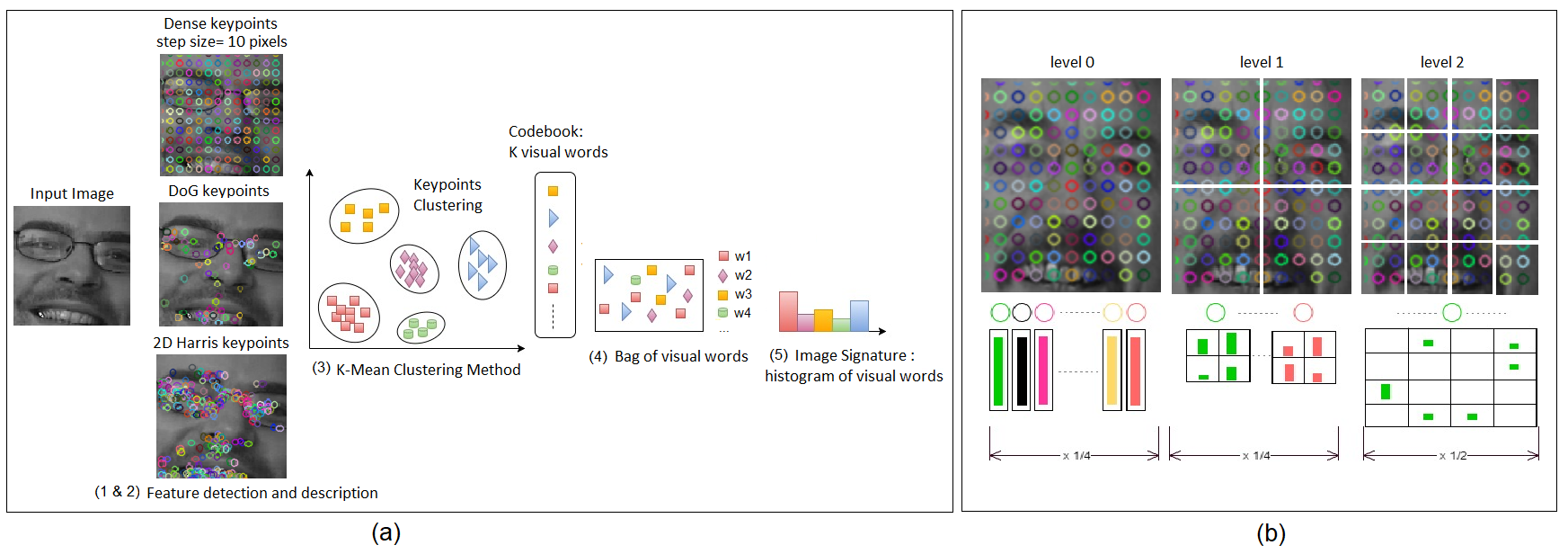}
\caption{(a): Standard BoVW representation for facial expressive image. (b): Three level spatial pyramid example.}
\label{classicalBoVW}
\end{figure*}

The standard steps for deriving the signature for a facial expression image using BoVW are represented in figure \ref{classicalBoVW} (a), (1): keypoints localization from the image, (2): keypoints description using local descriptors, (3): vector quantization for the descriptors by clustering them into $k$-clusters, using clustering methods, resulting a visual words vocabulary which forms the codebook, (4): establishing the signature of each image by accumulating the visual words into a histogram, (5): normalizing the histogram by dividing the frequency of each visual word over the total number of visual words, and (6): training a classifier using the obtained image signatures for classification task. 
 
BoVW model has been the most frequent and dominant used technique for visual content description. However this approach has some drawbacks that affect the performance. First, during the feature detection process, a large number of keypoints are located. This increases the computational process, in addition to the fact that most of these keypoints arise in the background regions. The second problem is the poor clustering, when the usual clustering method is Lloyd's algorithm referred as k-means algorithm, due to the fact that several local features are encoded with the same visual word. The third limitation is that as standard BoVW represents an image as an unordered collection of local descriptors, thus the spatial organization of the data is lost. The final drawback is the weighting scheme, where standard BoVW considers all visual words equally while there might be some visual words that are of greater importance. 

In the literature, many attempts have been conducted to improve standard BoVW model. In \cite{1641019}, an extension of spatial pyramid matching is proposed to exploit the spatial information. This technique works by partitioning the image into increasingly fine sub-regions and computing histograms of local features found inside each sub-region. The technique shows significantly improved performance on scene categorization tasks. In \cite{zhang2011generating}, descriptive visual words and descriptive visual phrases are proposed as visual correspondences to text words and phrases, where visual phrases refer to the frequently co-occurring visual word pairs. In \cite{xie2013weighted}, the descriptive ability of visual vocabulary has been investigated by proposing a weighting based method. In \cite{altintakan2015towards}, k-means clustering method has been replaced by self-organizing maps (SOM) in codebook generation as an alternative method. Obviously, state-of-the-art methods proposed to improve the standard BoVW method are dealing with one limitation at a time. In this paper, we perform several improvements, almost at each step of the standard method.

Our main contributions are: first, we investigate the use of BoVW for acted and spontaneous facial expression recognition. More specifically, we search for the best feature detector that suits our application. We integrate the use of k-means++ method \cite{arthur2007k} with BoVW method instead of k-means algorithm in an attempt to obtain a more distinctive codebook. We introduce relative conjunction matrix in order to preserve the spatial organization of the data. We introduce an efficient weighting scheme based on term-frequency inverse-document-frequency (TF-IDF), in order to scale up the rare visual words while damping the effect of the frequent visual words. Finally, for learning distinctive classifiers, we use the histogram intersection kernel since we experience much faster training than with the RBF kernel used by Support Vector Machine (SVM). Our choice for using BoVW method as a technique to tackle facial expression classification is motivated by the fact that differences between facial expressions are contained in the changes of location, shape and texture of facial clues (eyes, nose, eyebrows, etc.). We also want to explore the generalization power of geometrical based methods in case of strong geometrical deformations on faces (which is the case of acted emotions) and in case of more subtle deformations (which is the case of spontaneous facial expressions).

\section{\uppercase{Beyond Standard Bag of Visual Words Model}}
\label{SBOVW}

\subsection{Feature Selection and Description}

\noindent In image classification, low level visual features are used to represent different geometrical properties. Selecting these features plays a key factor in developing effective classification. In order to recognize facial expressions, low level visual features could be extracted from the facial deformations in the geometry of the facial shape. For example, anger on a face can be characterized by: eyebrows pulled down, upper lids pulled up, lower lids pulled up, lips may be tightened. Thereby, facial visual clues such as: eyes, nose, mouth, cheeks, eyebrow, forehead, etc. (Region of Interest: RoI) provide observable changes when an emotion occurs. Therefore, a feature detector that locates keypoints around those RoI would limit the risk of generating a huge number of redundant keypoints. Back to figure \ref{classicalBoVW} (a), we can see that 2D-Harris detector is focused on locating keypoints over the RoI. Although DoG detector has also focused on RoI, the keypoints are not as numerous as required. And thought dense feature extraction is known to be good for many classification problems \cite{furuya2009dense}, for facial expression recognition, the located keypoints are huge and redundant.

 Then, the extracted keypoints are described using local descriptors, in our case SIFT descriptors because they are invariant to image transformations, lighting variations and occlusions while being rich enough to carry enough discriminative information.

\subsection{k-means++ Clustering Algorithm}

\noindent The next step is to perform vector quantization, in order to quantize the space into a  discrete number of visual words. This step is important to map the image from a set of high-dimensional descriptors to a list of visual word numbers and though to provide a distinctive codebook. The usual method is to use k-means method. In most of the time, simple k-means algorithm generates arbitrary bad clustering specially when it is unbounded between n-data points and k-integers (pre-defined number of clusters) \cite{arthur2007k}. The simplicity of the k-means algorithm comes at the price of accuracy. To tackle this problem, we propose the use of k-means++ algorithm which is the result of augmenting k-means algorithm with a randomized seeding technique. The augmentation improves both the speed and the accuracy of k-means. The main steps of k-means++ clustering are:

\begin{enumerate}
    \item Choose an initial center uniformly at random from the data points.
    \item For each data point $x$, compute $D(x)$, the distance between x and the nearest center that has already been chosen.
    
    \item Choose one new data point at random as a new center, using a weighted probability distribution where a point $x$ is chosen with probability proportional to $D(x)^{2}$.
    
    \item Repeat steps 2 and 3 until a total of $k$ centers has been selected.
    
    \item Proceed as with standard k-means algorithm.
\end{enumerate}

\subsection{Relative Conjunction Matrix}
\noindent The BoVW approach describes an image as a bag of discrete visual words. The frequency distributions of these words are used for image categorization. The standard BoVW approach yields to a not complete representation of the data due to the fact that image features are modeled as independent and orderless visual words.  Thus there is no explicit use of visual word positions within the image. Traditional visual words based methods suffer when faced with similar appearances but distinct semantic concepts \cite{aldavert2015study}. In this study, we assume that establishing spatial dependencies might be useful for preserving the spatial organization of data. Thus we develop a novel facial image representation which uses the concept of the relative conjunction matrix to take into account links between the visual words. 

A relative conjunction matrix of visual words defines the spatial order by quantifying the relationships of each visual word with other visual words. To establish the correlation between the visual words, the neighborhood of each visual word feature is used. Thereby, a facial image is described by a histogram of pair-wise visual words. It provides a more discriminative representation since it contains the spatial arrangement of the visual words.

We define a relative conjunction matrix to establish pairs of visual words by looking for all possible pairs of visual words. This can be considered as a representation of the contextual distribution of each visual word with respect to other visual words of the vocabulary. The relative conjunction matrix $C$ has a size $N \times N$, where $N$ is the vocabulary size. Each element $C_{i,j}$ represents the pair of one independent feature to another. The obtained $C$ has all possible pairs. Each row vector of $C$ stores how many times a particular visual word (for example $W_{1}$) occurs  with any other visual words (for example $W_{2}$, $W_{3}$, $W_{4}$, ..., $W_{N}$). For a particular facial expression, if any two visual words have similar contextual distribution, that means they are capturing something similar. Thus, they are related to each other. The diagonal and the upper part of $C$ are considered for quantification. For quantifying this new representation, we adopt the method used in \cite{scovanner20073}, in which the correlation between the distribution vectors of any two visual words is computed. If the correlation is above a certain threshold (experimentally we found that $0.6$ is a good threshold), we join them together and their corresponding frequency counts from their initial histogram into a new grouping histogram.

\subsection{TF.IDF weighting scheme}

\noindent Weighting of visual words is crucial for classification performance but standard BoVW just normalizes the visual words by dividing them with the total number of visual words in the image. In \cite{van2010visual}, the authors investigate several types of soft-assignment of visual words to image features. They prove the fact that choosing the right weight scheme can improve the recognition performance. Each weight has to take into account the importance of each visual word in the image. For facial expression recognition, we are interested in scaling up the weights corresponding to visual words extracted from the nose, the eyebrows, and the mouth etc. while damping the effect of frequent visual words that describe the hair and some non-deformable regions like the background. Therefore, it is possible to leverage the usage of term- frequency inverse-document-frequency (TF.IDF) \cite{leskovec2014mining} weighting scheme to scale up the rare visual words while scaling down the frequent ones. 

 The standard weighting method used in traditional BoVW approach is equivalent to the term frequency referred as $TF(vw)$, where $vw$ is the visual word. It measures how frequently a visual word occurs in an image. It is normalized by dividing it with the total number of visual words in the image. The utilization of $TF(vw)$ in classification is rather straightforward and usually results in decreased accuracy due to the fact that all visual words are considered equally important. 

 However, inverse-document-frequency referred as $IDF(vw)$ assigns different weights to features. It provides information about the general distribution of visual word $vw$ amongst facial images of all classes. The utilization of $IDF(vw)$ is based on its ability to distinguish between visual words with some semantical meanings and simple visual words. The $IDF(vw)$ measures how unique a $vw$ is and how infrequently it occurs across all training facial expression images.

\[IDF(vw) = \log(\frac{T}{n_{vw}})\]

\noindent $T$: total number of training images. \\
$n_{vw}$: number of occurrences of $vw$ in the whole training database $T$.

However, if we assume that certain visual words may appear a lot of times but have little importance, then we need to weight down the most frequent visual words while scaling up the rare ones, by computing the term-frequency-inverse-document-frequency referred as $TF.IDF$. 
\noindent Where:

\begin{equation}
TF.IDF(vw)= TF(vw) \cdot \log(\frac{T}{n_{vw}})
    \label{tfidf}
\end{equation}


 The $\mbox{TF.IDF}_{vw,I}$ assigns to visual word $vw$ a weight in image $I$, where $I \in T$, such that: high weight when $vw$ occurs frequently within a small number of images, thus lending high discriminating power to those images. Low weight when the $vw$ occurs less frequently in an image, or occurs in many images (for example, $vw \in$ background), thus offering a less pronounced relevance signal.

    
    

\section{\uppercase{Facial Expression Classification}}
\label{WCBOVW}

\noindent The proposed Improved Bag-of-Visual-Word model (ImpBoVW) for facial expression classification is summarized as follows: 


\begin{enumerate}
     \item Locate and extract salient features (keypoints) from facial images either based on a feature detector such as: Difference of Gaussian (DoG), 2D-Harris detector, or by defining a grid with pre-specified spatial step (for example 5 pixels) to extract local feature descriptors from.
     
     \item Describe local features over the selected salient keypoints, the SIFT descriptor is used. 
     
     \item Quantize the descriptors gathered from all the keypoints by clustering them into $k$-clusters, using k-Mean++. It quantizes the space into a pre-specified number (vocabulary size) of visual words. The cluster centers represent the visual words. Resulting visual words vocabulary forms the codebook.
     
      \item Map a set of high dimensional descriptors into a list of visual words by assigning the nearest visual word to each of its features in the feature space. This results the histogram of visual words. It summarizes the entire facial image and it is considered as the signature of the image. 

     \item Build feature grouping among the words. A co-occurrence based criterion is used for learning discriminative word groupings using Relative Conjunction Matrix.
    
     \item Introduce the proper $TF.IDF$ weighting scheme based on equation \ref{tfidf}.
    
     \item Train a SVM classifier over the diagonal and the upper parts of the weighted conjunction matrix for facial expressions recognition. Histogram Intersection kernel (equation \ref{histointer}) is used by SVM to learn a discriminative classifier. 
    
\end{enumerate}

\section{\uppercase{BoVW Model with Spatial Pyramid Representation}}
\label{SP}

\noindent In order to evaluate the effectiveness of the proposed method and for fair comparison, we have also implemented the spatial pyramid BoVW model presented in \cite{1641019} in addition to the standard BoVW method. BoVW method with spatial pyramid has shown significantly improved performance on scene categorization tasks \cite{zhu2016bag}. Spatial Pyramid BoVW (SP BoVW) representation is an extension of an orderless BoVW image representation. It aims at subdividing the image into increasingly fine resolutions and at computing histograms of local features. Thus, it aggregates statistics of local features over fixed sub-regions. A match between two keypoints occurs if they fall into the same cell of the grid. Suppose $X$ and $Y$ are two sets of vectors in a $d$-dimensional feature space. Let us construct a sequence of grids at resolutions $0,...,L$, such that the grid at level $l$ has $2^{l}$ cells along each dimension, for a total of $D =2^{dl}$ cells. Let $H^{l}_{X}$ and $H^{l}_{Y}$ denote the histograms of $X$ and $Y$ at this resolution, such that $H^{l}_{X}(i)$ and $H^{l}_{Y}(i)$ are the number of points from $X$ and $Y$ that fall into the $i$th cell of the grid. Then the number of matches at level $l$ is given by the histogram intersection function:

\begin{equation}
    I(H^{l}_{X},H^{l}_{Y}) = \sum_{i=1}^{D}min(H^{l}_{X}(i),H^{l}_{Y}(i))
    \label{histointer}
\end{equation}

\noindent The weight associated with level $l$ is proportional to the cell width at that level: $\frac{1}{2^{L-l}}$.

\noindent However, pyramid match kernel (PMK) aims at penalizing matches found in larger cells since they involve increasing dissimilar features, thus:

\begin{equation}
\begin{aligned}
 PMK^L(X,Y) &= I(H^{l}_{X},H^{l}_{Y})^L + \sum_{l=0}^{L-1}\frac{1}{2^{L-l}}(I^{l}-I^{l+1}) \\
& = \frac{1}{2^{L}}I^{0} + \sum_{l=1}^{L}\frac{1}{2^{L-l+1}}I^{l}
\end{aligned}
 \label{mercerkern}
\end{equation}


\noindent Equation \ref{mercerkern} is known as Mercer kernel that combines both the histogram intersection and the pyramid match kernel \cite{grauman2007pyramid}.

\noindent For spatial pyramid representation, the pyramid matching in $2D$-image space is performed and k-means clustering algorithm is used to quantize all feature vectors into M discrete channels. Each channel gives two dimensional vectors, $X_{m}$ and $Y_{m}$ corresponding to the coordinates of the features of channel m found in the respective images.

\noindent The final kernel (FK) represents the sum of the separate channel kernels:

\begin{equation}
\mathcal{FK}^{L}(X,Y) = \sum_{m=1}^{M} PMK^{L}(X_{m},Y_{m})
\end{equation}

\noindent Figure \ref{classicalBoVW} (b) represents the construction of a three level spatial pyramid. The image has different feature types, indicated by different colors. At the top, the facial image is sliced at two different levels of resolution. Then, for each resolution and each channel, the features that fall in each spatial bin are counted. Finally, each spatial histogram is weighted according to equation \ref{mercerkern}.


\section{\uppercase{Experimental Results}}

\noindent In this section, we present the experimental design used to evaluate the proposed algorithm and compare it to other approaches. First, we present the datasets and protocols. Then, we describe the evaluation procedure and finally we present the results.

\subsection{Data Exploration}

\noindent For effective and fair comparison, four different databases are used: three with Ekman's caricatured facial expressions (the JAFFE database \cite{27}, the extended Cohn Kanade database (CK+) \cite{kanade2000comprehensive} and the MMI facial expression databases \cite{pantic2005web}) and one with spontaneous expressions (the DynEmo database \cite{28}). 


 \textbf{The JAFFE Database}: it is a well-known database made of acted facial expressions. It contains 213  facial expression images with 10 different identities. It includes: \enquote{happy}, \enquote{anger}, \enquote{sadness}, \enquote{surprise}, \enquote{disgust}, \enquote{fear} and \enquote{neutral}. The head is in frontal pose. The number of images corresponding to each is roughly the same (around 21). Seven identities are used during the training phase while three other identities are used during the test phase.

 \textbf{The CK+ Database}: it is a widely used database containing acted Ekman's expressions. It is composed of 123 different identities. In our study, we pick out the last frame which represents the peak of emotion. Each subject performed different sessions corresponding to different emotions. In total we collected 306 images associated with its ground truth label.

 \textbf{The MMI  Database}: it is composed of more than 1500 samples of both static images and image sequences of faces in frontal and in profile views displaying various facial expressions. It is performed by 19 different people both genders, ranging in age between 19 and 62, having a different ethnic background. Images are given a single label that belongs to one of six Ekman emotion. In this study, we collect 600 static frontal images and we exploit the 900 sequences to extract from each sequence other static expressive images. From this database we create a collection of 1900 static images. This database is used to check the scalability of the proposed approach on a large databse.

 \textbf{The DynEmo Database}: it is a database containing elicited facial expressions. It is made of six spontaneous expressions which are: \enquote{irritation}, \enquote{curiosity}, \enquote{happiness}, \enquote{worried}, \enquote{astonishment}, and \enquote{fear}. The database contains a set of 125 recordings of facial expressions of ordinary Caucasian people (ages 25 to 65, 182 females and 176 males) filmed in natural but standardized conditions. 480 expressive images that correspond to 65 different identities are extracted from the database. The head is not totally in frontal pose. The number of images corresponding to each of the six categories of expressions is roughly the same (80 images per class). The dataset is challenging since it is closer to natural human behaviour and each person has a different way to react to a given emotion. 



\textbf{\emph{Training Protocol}}: Identities that appear in the training sets do not appear in the test sets.

\noindent \textbf{\emph{Train set}}: 70\% of randomly shuffled images per class are picked out as training sets. Therefore, for the JAFFE dataset we have 143 training images (20 images per class), for the CK+ dataset we have 216 training images (36 images per class), for the MMI dataset we have 1330 images (around 221 images per class) and finally for the DynEmo we have 360 training images (60 images per class). 

\noindent \textbf{\emph{Development set}}: Leave-one-out cross validation is considered over the training set to tune the algorithm hyper-parameters.

\noindent \textbf{\emph{Test set}}: 30\% images per class are picked out as test set, that is 70 test images from the JAFFE set (10 images per class), 90 test images (15 images per class) for the CK+ set, 570 test images (95 images per class) for the MMI set and 120 test images from the DynEmo set (20 images per class), randomly shuffled, to test the performance of the proposed method.

\subsection{Experimental setup and Results}
\label{expdes}

\noindent We focus the experimental evaluation of the proposed method on the following four questions:\textit{ What is the best feature detector that locates salient and reliable feature points for facial expression recognition? Does each of the proposed novelties improve the performance of the SBoVW? What is the influence of using k-means++? Is the proposed approach efficient for facial recognition and scalable for larger databases?}

 The proposed model has many parameters that influence its classification performance: the usage of weighting scheme $TF.IDF$, the usage of Relative Conjunction Matrix (RCM), the combination of $TF.IDF$ weighting scheme along with RCM, the usage of K-mean and K-mean++ as clustering methods, the choice of the best feature detector and descriptor. Thereby, in order to answer the first three questions, we report performance of facial expression recognition with SBoVW representation along with each novelty. The JAFFE (caricatured facial expressions) and the DynEmo (non-caricatured facial expressions) databases are used for the method evaluation and setting and to figure out the best feature detector that suit facial expression recognition. The final question is addressed using the CK+ and the MMI databases to establish the performance of ImpBoVW model and to compare it with  SBoVW and SP BoVW.

 Multi-class classification is done using SVM trained using one-versus-all rule. The histogram intersection kernel presented in equation \ref{histointer} is used. Compared to RBF kernel, we experience faster computation while accuracy rate has a smaller variance. One-hold-out cross-validation method is used over the training set in order to tune the algorithm hyper-parameters such as the regularization parameter C ( the optimal value is $10.0$), the gamma parameter which stands for Gaussian kernel to handle non-linear classification if considered. We fix the vocabulary size to 2000 visual words, since experimentally it shows the best classification performance. For spatial pyramid representation, we notice that at level-2 and level-3 same performance is achieved. Thus to reduce the complexity of feature computation, we only consider two levels. 

\begin{figure}[]
\centering
\includegraphics[width=3 in]{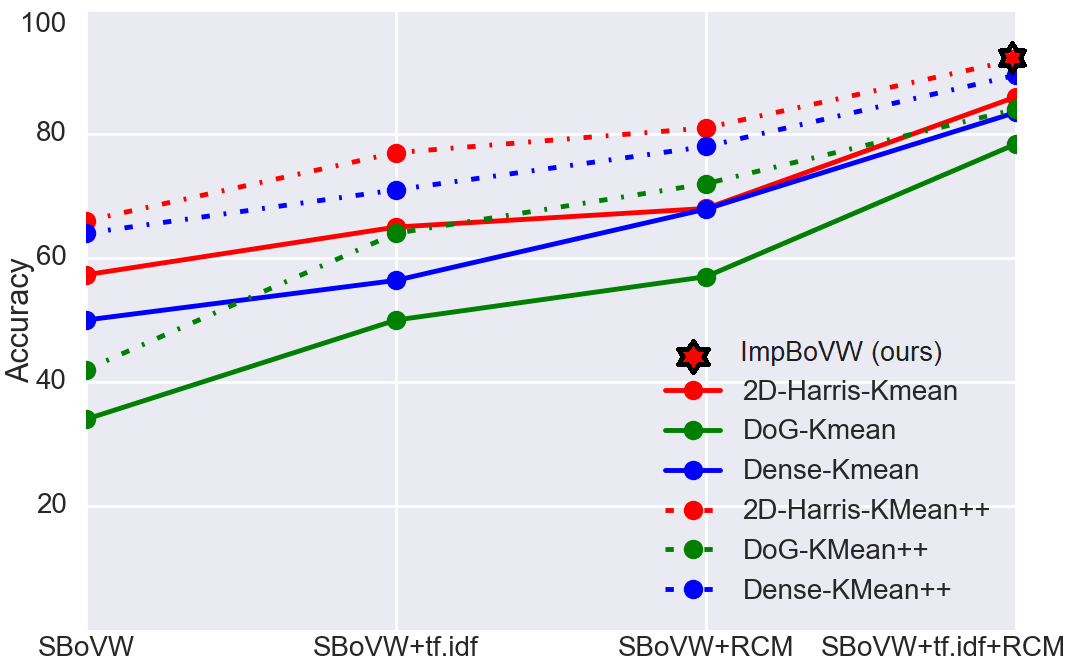}
\caption{Classification accuracy obtained for SBoVW associated with improved novelties over acted expressions (JAFFE) combined with different feature detection methods.}
\label{evaluationJaffe}
\end{figure}

\begin{figure}[]
\centering
\includegraphics[width=3 in]{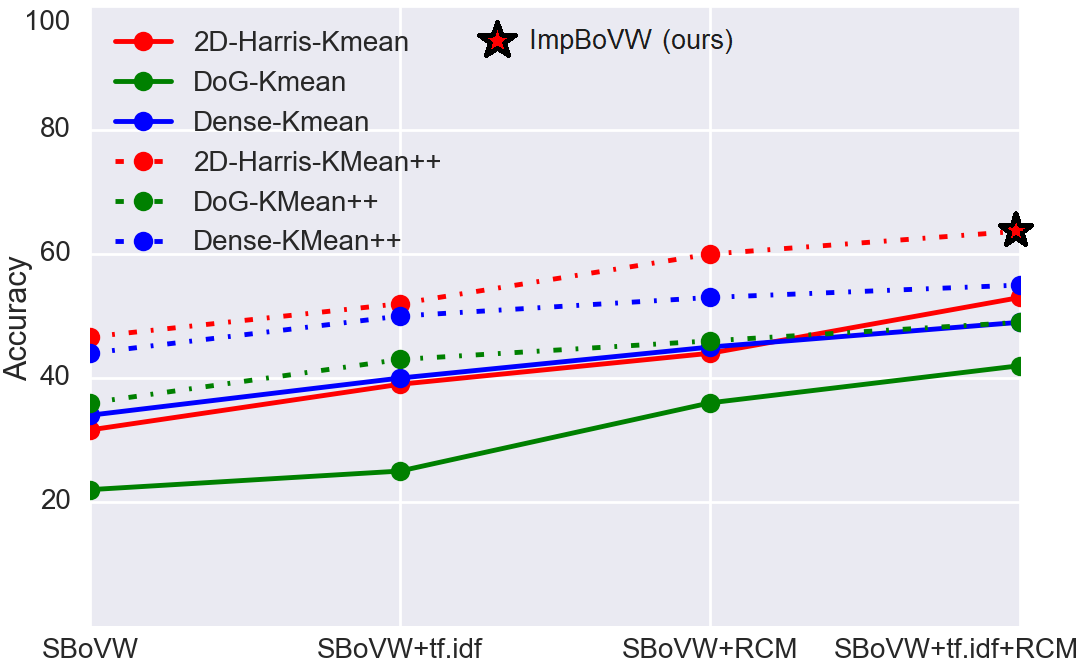}
\caption{Classification accuracy obtained for SBoVW associated with improved novelties over spontaneous expressions (DynEmo) combined with different feature detection methods.}
\label{evaluationDynEmo}
\end{figure}

 Figure \ref{evaluationJaffe} represent the performance of each novelty and its contribution to the final ImpBoVW over the JAFFE database. ImpBoVW represents the best final model which is a combination of SBoVW representation along with RCM and $TF.IDF$ weighting scheme (representd as a star in the figure \ref{evaluationJaffe} and \ref{evaluationDynEmo}). Its quantization is based on K-mean++. Its features are located using 2D-Harris detector and described using SIFT descriptor. The final average recognition rate obtained over the JAFFE database is 92\%. If we compare ImpBoVW model using DoG feature detector, we got 84\% accuracy while 89,5\% is achieved using dense features. 2D-Harris is less computationally expensive than dense features due to the fact that it produces less redunadat features. In addition, figure \ref{evaluationJaffe} shows that if each novelty stand alone along with SBoVW, a noticed increase in the recognition rate is achieved. More importantly, the figure shows that the usage of K-means++ increases the classification rate significantly.

 In order to fairly estimate the performance of ImpBoVW model and to get a fair judgment about the choice and the best feature detector and clustering method, we evaluate the same steps over the DynEmo database. DynEmo represents naturalistic static images where subtle deformations occur. ImpBoVW model using 2D-Harris detector achieved 64\% recognition rate (see figure \ref{evaluationDynEmo}), 55\% using dense features and 49\% using DoG features, all along with K-means++. However, for the same setting using K-means, the following results achieved respectively: 53\% (2D-Harris), 49\% (dense)  and 42\% (DoG) respectively. Figures \ref{evaluationJaffe} and \ref{evaluationDynEmo} prove the fact that: 2D-Harris is a good detector for localizing salient features, K-means++ clustering method has a significant contribution over the final recognition rate, the combination of SBoVW along with RCM and $TF.IDF$ improves the representational quality of the image signature.

\begin{figure}[]
\centering
\includegraphics[width=3 in]{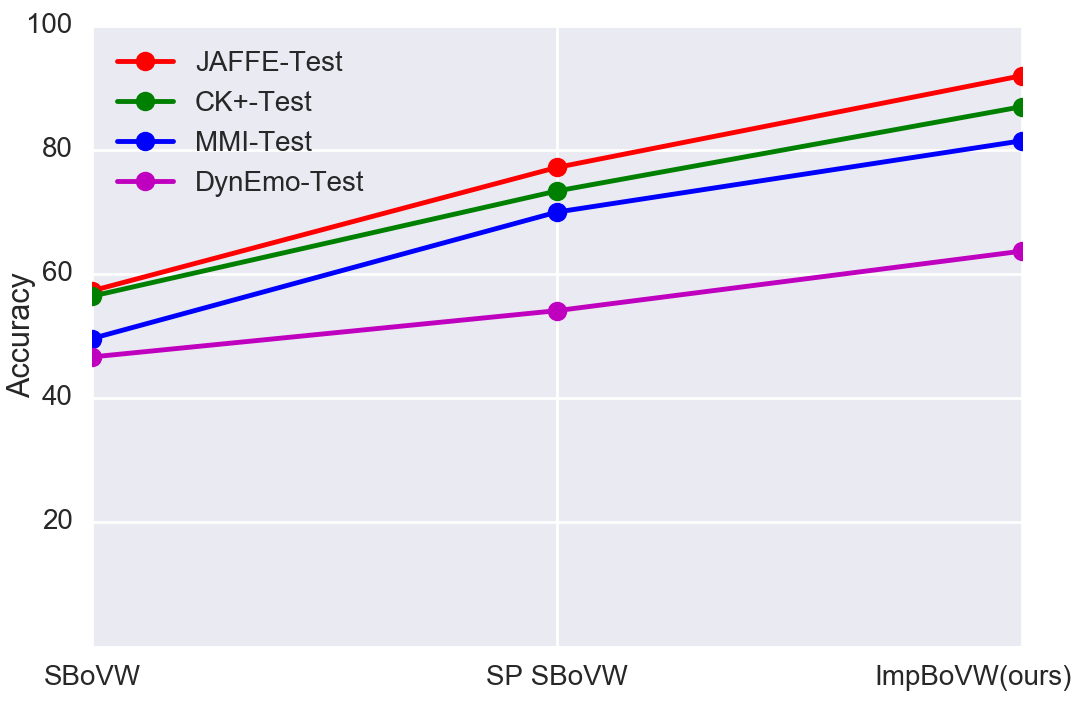}
\caption{Performance over four databases compared with SBoVW and SP BoVW.}
\label{resulttable}
\end{figure}

\begin{figure}[]
\centering
\includegraphics[width=3.2 in]{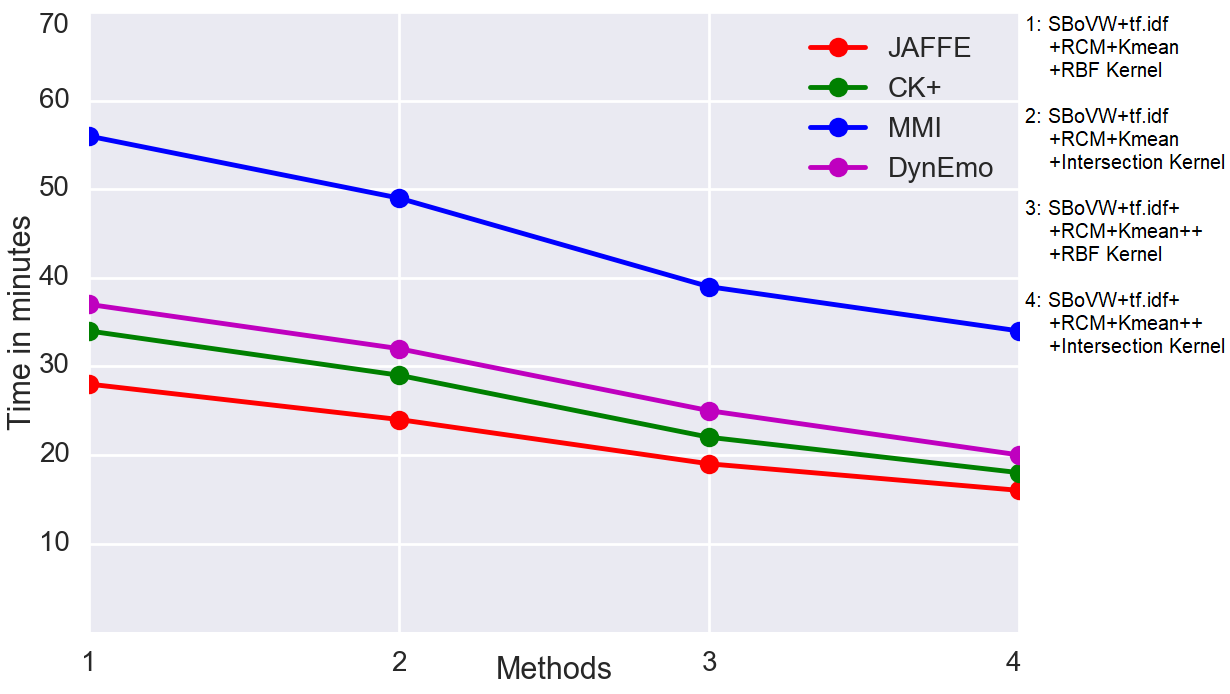}
\caption{Computational time for generating the BoVW features.}
\label{timecomplexity}
\end{figure}

\textbf{\emph{Computational Time Performance}}: In order to compare the time complexity of the proposed method with k-means++ and the histogram intersection kernel, we report in figure \ref{timecomplexity} the time taken in minutes for the whole training phase of SBoVW+RCM+TF.IDF with either Kmean or Kmean++ and with either RBF kernel or Intersection kernel. Method number 4 represent ImpBoVW. The contribution of k-Mean++ helps speeding up the process of vector quantization. In addition, the histogram Intersection kernel has also contributed in decreasing the complexity of the learning part by SVM. Figure \ref{timecomplexity} show that our method achieved a good compuational time compared to the original algorithm.

\section{\uppercase{Conclusion}}
\noindent In this paper, we introduced an improved BoVW approach for automatic emotion recognition. We examined several aspects of the SBoVW approach that are linked directly to gain classification performance, speed and scalability. It has been proved that Harris detector is suitable for emotion recognition. It selects adaptable and reliable salient keypoints. We improve the codebook generation process through employing k-means++ as a clustering method, wherein we gain speed and accuracy. The importance of spatial organization of the data has been examined, and we optimized the feature representation by introducing a relative conjunction matrix to preserve the spatial order. We properly weighted the visual words after preserving the spatial order using TF.IDF based on their occurrences. Histogram intersection kernel has been used to decrease the complexity of the algorithm. We implemented SP BoVW for comparison purpose and different feature detection methods are evaluated. We noticed that the geometrical based method is robust if strong geometrical deformations are present on the face, which is the case with acted expressions. However for spontaneous expressions where the facial deformations are more subtle, it appears that geometrical based methods alone are not so efficient to achieve good performance due to the fact that each person has a different way to react to a given emotion. For future work, the idea would be to combine the proposed approach with an appearance based facial expression recognition method we developed in \cite{visapp17}, in order to take benefit of the advantages of both approaches.

\bibliographystyle{apalike}

{\small


\begin{thebibliography}{}

\bibitem[Aldavert et~al., 2015]{aldavert2015study}
Aldavert, D., Rusi{\~n}ol, M., Toledo, R., and Llad{\'o}s, J. (2015).
\newblock A study of bag-of-visual-words representations for handwritten
  keyword spotting.
\newblock {\em International Journal on Document Analysis and Recognition
  (IJDAR)}, 18(3):223--234.

\bibitem[Altintakan and Yazici, 2015]{altintakan2015towards}
Altintakan, U.~L. and Yazici, A. (2015).
\newblock Towards effective image classification using class-specific codebooks
  and distinctive local features.
\newblock {\em IEEE Transactions on Multimedia}, 17(3):323--332.

\bibitem[Arthur and Vassilvitskii, 2007]{arthur2007k}
Arthur, D. and Vassilvitskii, S. (2007).
\newblock k-means++: The advantages of careful seeding.
\newblock In {\em Proceedings of the eighteenth annual ACM-SIAM symposium on
  Discrete algorithms}, pages 1027--1035. Society for Industrial and Applied
  Mathematics.

\bibitem[Chanti and Caplier, 2017]{visapp17}
Chanti, D.~A. and Caplier, A. (2017).
\newblock Spontaneous facial expression recognition using sparse
  representation.
\newblock In {\em Proceedings of the 12th International Joint Conference on
  Computer Vision, Imaging and Computer Graphics Theory and Applications
  (VISIGRAPP 2017)}, pages 64--74.

\bibitem[Furuya and Ohbuchi, 2009]{furuya2009dense}
Furuya, T. and Ohbuchi, R. (2009).
\newblock Dense sampling and fast encoding for 3d model retrieval using
  bag-of-visual features.
\newblock In {\em Proceedings of the ACM international conference on image and
  video retrieval}, page~26. ACM.

\bibitem[Grauman and Darrell, 2007]{grauman2007pyramid}
Grauman, K. and Darrell, T. (2007).
\newblock The pyramid match kernel: Efficient learning with sets of features.
\newblock {\em Journal of Machine Learning Research}, 8(Apr):725--760.

\bibitem[Hariri et~al., 2017]{hariri2017geometrical}
Hariri, W., Tabia, H., Farah, N., Declercq, D., and Benouareth, A. (2017).
\newblock Geometrical and visual feature quantization for 3d face recognition.
\newblock In {\em VISAPP 2017 12th Joint Conference on Computer Vision, Imaging
  and Computer Graphics Theory and Applications}.

\bibitem[Ionescu et~al., 2013]{ionescu2013local}
Ionescu, R.~T., Popescu, M., and Grozea, C. (2013).
\newblock Local learning to improve bag of visual words model for facial
  expression recognition.
\newblock In {\em Workshop on challenges in representation learning, ICML}.

\bibitem[Kanade et~al., 2000]{kanade2000comprehensive}
Kanade, T., Cohn, J.~F., and Tian, Y. (2000).
\newblock Comprehensive database for facial expression analysis.
\newblock In {\em Automatic Face and Gesture Recognition, 2000. Proceedings.
  Fourth IEEE International Conference on}, pages 46--53. IEEE.

\bibitem[Lazebnik et~al., 2006]{1641019}
Lazebnik, S., Schmid, C., and Ponce, J. (2006).
\newblock Beyond bags of features: Spatial pyramid matching for recognizing
  natural scene categories.
\newblock In {\em 2006 IEEE Computer Society Conference on Computer Vision and
  Pattern Recognition (CVPR'06)}, volume~2, pages 2169--2178.

\bibitem[Leskovec et~al., 2014]{leskovec2014mining}
Leskovec, J., Rajaraman, A., and Ullman, J.~D. (2014).
\newblock {\em Mining of massive datasets}.
\newblock Cambridge University Press.

\bibitem[Lyons et~al., 1998]{27}
Lyons, M., Akamatsu, S., Kamachi, M., and Gyoba, J. (1998).
\newblock Coding facial expressions with gabor wavelets.
\newblock In {\em Automatic Face and Gesture Recognition, 1998. Proceedings.
  Third IEEE International Conference on}, pages 200--205. IEEE.

\bibitem[Pantic et~al., 2005]{pantic2005web}
Pantic, M., Valstar, M., Rademaker, R., and Maat, L. (2005).
\newblock Web-based database for facial expression analysis.
\newblock In {\em Multimedia and Expo, 2005. ICME 2005. IEEE International
  Conference on}, pages 5--pp. IEEE.

\bibitem[Peng et~al., 2016]{peng2016bag}
Peng, X., Wang, L., Wang, X., and Qiao, Y. (2016).
\newblock Bag of visual words and fusion methods for action recognition:
  Comprehensive study and good practice.
\newblock {\em Computer Vision and Image Understanding}, 150:109--125.

\bibitem[Scovanner et~al., 2007]{scovanner20073}
Scovanner, P., Ali, S., and Shah, M. (2007).
\newblock A 3-dimensional sift descriptor and its application to action
  recognition.
\newblock In {\em Proceedings of the 15th ACM international conference on
  Multimedia}, pages 357--360. ACM.

\bibitem[Sivic and Zisserman, 2003]{sivic2003video}
Sivic, J. and Zisserman, A. (2003).
\newblock Video google: A text retrieval approach to object matching in videos.
\newblock In {\em null}, page 1470. IEEE.

\bibitem[Tcherkassof et~al., 2013]{28}
Tcherkassof, A., Dupr{\'e}, D., Meillon, B., Mandran, N., Dubois, M., and Adam,
  J.-M. (2013).
\newblock Dynemo: A video database of natural facial expressions of emotions.
\newblock {\em The International Journal of Multimedia \& Its Applications},
  5(5):61--80.

\bibitem[Van~Gemert et~al., 2010]{van2010visual}
Van~Gemert, J.~C., Veenman, C.~J., Smeulders, A.~W., and Geusebroek, J.-M.
  (2010).
\newblock Visual word ambiguity.
\newblock {\em IEEE transactions on pattern analysis and machine intelligence},
  32(7):1271--1283.

\bibitem[Xie et~al., 2013]{xie2013weighted}
Xie, Y., Jiang, S., and Huang, Q. (2013).
\newblock Weighted visual vocabulary to balance the descriptive ability on
  general dataset.
\newblock {\em Neurocomputing}, 119:478--488.

\bibitem[Zhang et~al., 2011]{zhang2011generating}
Zhang, S., Tian, Q., Hua, G., Huang, Q., and Gao, W. (2011).
\newblock Generating descriptive visual words and visual phrases for
  large-scale image applications.
\newblock {\em IEEE Transactions on Image Processing}, 20(9):2664--2677.

\bibitem[Zhu et~al., 2016]{zhu2016bag}
Zhu, Q., Zhong, Y., Zhao, B., Xia, G.-S., and Zhang, L. (2016).
\newblock Bag-of-visual-words scene classifier with local and global features
  for high spatial resolution remote sensing imagery.
\newblock {\em IEEE Geoscience and Remote Sensing Letters}, 13(6):747--751.

\end{thebibliography}
}

\vfill
\end{document}